\documentclass{article}

\usepackage{arxiv}

\usepackage{amsmath,amsfonts,bm}

\def\eqref#1{equation~\ref{#1}}

\def\1{\bm{1}}

\DeclareMathAlphabet{\mathsfit}{\encodingdefault}{\sfdefault}{m}{sl}
\SetMathAlphabet{\mathsfit}{bold}{\encodingdefault}{\sfdefault}{bx}{n}

\usepackage[utf8]{inputenc} 
\usepackage[T1]{fontenc}    
\usepackage{hyperref}       
\usepackage{url}            
\usepackage{booktabs}       
\usepackage{amsfonts}       
\usepackage{nicefrac}       
\usepackage{microtype}      
\usepackage{lipsum}
\usepackage{graphicx}
\usepackage[table]{xcolor}
\usepackage{multirow}
\usepackage{longtable}
\graphicspath{ {./images/} }

\title{PLanTS: Periodicity-aware Latent-state Representation Learning for Multivariate Time Series}

\author{
 Jia Wang \\
  Department of Computer Science\\
  Indiana University\\
  Bloomington, IN 47408 \\
  \texttt{jw316@iu.edu} \\
   \And
Xiao Wang \\
  Department of Computer Science\\
  Indiana University\\
  Bloomington, IN 47408 \\
  \texttt{xw90@iu.edu} \\
  \And
 Chi Zhang \\
  Department of Biomedical Engineering\\
  Oregon Health \& Science University\\
  Portland, OR 97201 \\
  \texttt{zhangchi@ohsu.edu} \\
}

\begin{document}
\maketitle
\begin{abstract}
Multivariate time series (MTS) are ubiquitous in domains such as healthcare, climate science, and industrial monitoring, but their high dimensionality, limited labeled data, and non-stationary nature pose significant challenges for conventional machine learning methods. While recent self-supervised learning (SSL) approaches mitigate label scarcity by data augmentations or time point-based contrastive strategy, they neglect the intrinsic periodic structure of MTS and fail to capture the dynamic evolution of latent states.  We propose PLanTS, a periodicity-aware self-supervised learning framework that explicitly models irregular latent states and their transitions. We first designed a period-aware multi-granularity patching mechanism and a generalized contrastive loss to preserve both instance-level and state-level similarities across multiple temporal resolutions. To further capture temporal dynamics, we design a next-transition prediction pretext task that encourages representations to encode predictive information about future state evolution.  We evaluate PLanTS across a wide range of downstream tasks—including multi-class and multi-label classification, forecasting, trajectory tracking and anomaly detection. PLanTS consistently improves the representation quality over existing SSL methods and demonstrates superior runtime efficiency compared to DTW-based methods. Code is available at this repository: \href{https://github.com/JiaW6122/PLanTS}{https://github.com/JiaW6122/PLanTS}.
\end{abstract}


\section{Introduction}
Multivariate time series (MTS) data is now prevalent across a wide range of domains, including healthcare, climate science, and industrial monitoring\cite{zhang2018multivariate,nguyen2017review,cook2019anomaly}. However, MTS data is inherently high-dimensional, often non-stationary, and typically exhibit limited labeled instances, which presents significant challenges for supervised learning approaches\cite{montgomery2015introduction,cheng2015time,liu2022non}. In different application settings, tasks such as classification\cite{ismail2019deep}, forecasting\cite{lim2021time} and anomaly detection\cite{zamanzadeh2024deep} often require extracting distinct and task-specific information from the temporal signals. Training task-specific model for each objective is not only computational expensive but also lacks knowledge sharing across tasks. To overcome these limitations, self-supervised learning (SSL) has emerged as a promising paradigm for learning general-purpose representations from unlabeled time series data\cite{zhang2024self,trirat2024universal}.

Recent SSL methods typically rely on either handcrafted augmentations \cite{eldele2021time,zheng2024parametric} or context-based modeling \cite{yue2022ts2vec,fraikin2024t,lee2024soft} to construct positive and negative pairs for contrastive learning. These pairs are designed to encourage the model to learn representations that are invariant to noise and transformation, while preserving semantic similarity.

However, the effectiveness of representations on downstream tasks directly depends on the alignment between semantic similarity and the constructed pairwise relationships\cite{wang2022chaos,demirel2024unsupervised}. Real-world MTS datasets—such as electrocardiograms (ECGs), which capture cardiac condition of a patient—often show noisy quasi-periodic characteristic\cite{nagendra2011application,rhif2019wavelet}. Naive pairing strategies often neglect the underlying periodic structures, leading to false positive pairs (instances with different physiological phases being treated as similar) and false negative pairs (temporally shifted but semantically identical segments being treated as dissimilar). Such misalignment degrades the effectiveness of the contrastive objective and limits the downstream performance of the learned representations.

\begin{figure}[h]
\centering
\includegraphics[width=0.6\textwidth]{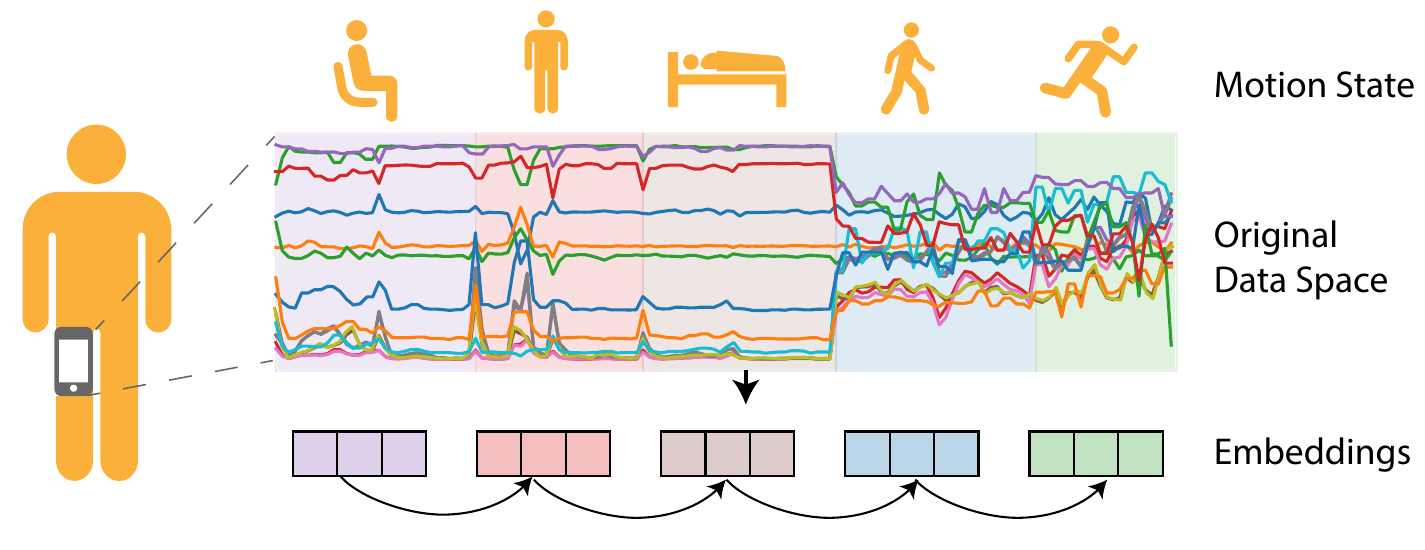}
\caption{Illustration of human activity recognition tasks using wearable sensors. Background colors in the original data space indicate ground-truth motion states. PLanTS is designed to not only distinguish between motion states but also model the underlying dynamic transitions in the latent space.}
\label{fig1}
\end{figure}

Furthermore, most existing SSL methods for time series focus on instance-wise or timestamp-wise contrastive learning at the raw data level\cite{yue2022ts2vec,fraikin2024t,lee2024soft}, without explicitly modeling the latent states and their temporal transitions. This is a critical limitation, as many real-world MTS applications involve non-stationary dynamics where the latent distribution of the signal evolves over time\cite{tonekaboni2021unsupervised}. For example, in Human Activity Recognition (HAR) tasks using wearable sensors, individuals transition gradually between distinct motion states (e.g., walking, sitting, running) with irregular and variable durations (Figure~\ref{fig1}). In such cases, learning representations that can not only discriminate between latent motion states but also capture the dynamics of state transitions is essential for accurately tracking and forecasting activity trajectories. Similarly, in critical care settings, identifying the patient's latent clinical state and modeling its evolution over time is key to understanding disease progression and informing timely treatment decisions\cite{schulam2015clustering,suresh2018learning}.

To address the above challenges, we propose PLanTS, a Periodicity-aware Latent-state representation learning framework for learning robust and generalizable representations from complex, non-stationary multivariate time series. To model irregular latent states, PLanTS introduces a multi-granularity generalized contrastive loss guided by periodicity. We hypothesize that dominant periodic patterns often correspond to transitions between latent states, and that latent state similarities should be modeled as continuous relations rather than binary positive/negative pairs.

In addition, PLanTS formulates a pretext task designed to capture the dynamic transitions between latent states. This task incorporates both latent state embeddings and timestamp information to explicitly model temporal dependencies across state transitions.

We conduct extensive experiments across a wide range of downstream tasks—including multi-class and multi-label classification, forecasting, trajectory
tracking and anomaly detection—using five publicly available MTS datasets. Notably, we evaluate on PTB-XL \cite{wagner2020ptb}, the largest publicly available clinical ECG waveform dataset to date, which contains 21,837 records from 18,885 patients. Results show that PLanTS consistently improves the representation quality over existing SSL methods and achieves state-of-the-art performance across diverse tasks.

The contributions of out work are as follows:
\begin{itemize}
    \item We propose PLanTS, a periodicity-aware, multi-granularity self-supervised learning framework for non-stationary multivariate time series.
    \item We introduce a generalized contrastive loss that leverages periodic similarity in the input space to learn invariant latent state representations, and design a temporal transition prediction task to model state evolution. 
    \item PLanTS outperforms existing state-of-the-art methods across five downstream tasks. Remarkably, it reduce the average MSE by 7.2\% and 9.1\% in ETTh1 and ETTm1; Achieve 64\% time saved than SoftCLT.
\end{itemize}

\section{Related Work}

\textbf{Self-supervised learning.} \: Self-supervised learning has emerged as a powerful paradigm for extracting informative representations from unlabeled data by formulating pretext tasks that transform unsupervised objectives into supervised learning problems\cite{liu2021self}. In natural language processing, common pretext tasks include next-token prediction and masked-token prediction\cite{devlin2019bert,rethmeier2023primer}, while in computer vision, tasks such as solving jigsaw puzzles\cite{noroozi2016unsupervised}, predicting image rotations\cite{gidaris2018unsupervised} and clustering augmented views\cite{caron2018deep} have been widely adopted. More recently, contrastive learning-exemplified by frameworks like SimCLR\cite{chen2020simple} and MoCo\cite{he2020momentum}-has gain significant attention by constructing multiple views of the same instance and encouraging alignment of positive pairs while pushing apart negative pairs based on InfoNCE loss\cite{oord2018representation}. However, many SSL methods developed for vision and language domains rely on domain-specific inductive biases-such as transformation invatiance and cropping invariance-that do not applicable to time series data, where periodic structures and temporal continuity are critical.

\textbf{Contrastive learning for time series data.}\: Recent studies have demonstrated the effectiveness of contrastive learning (CL) in time series representation learning. T-loss \cite{franceschi2019unsupervised} introduces a triplet-loss-based approach that employs time-based negative sampling for multivariate time series. TSTCC \cite{eldele2021time} proposes a temporal and contextual contrasting framework that generates two related views via weak and strong augmentations of the raw signal. TF-C \cite{zhang2022self} incorporates a time-frequency consistency mechanism to jointly learn time-domain and frequency-domain representations by aligning their respective local neighborhoods in latent space.
While these methods focus primarily on instance-level contrast, they often struggle with temporally-sensitive downstream tasks such as forecasting and anomaly detection. To address this, TS2Vec \cite{yue2022ts2vec} introduces a hierarchical contrastive strategy that combines instance-wise and temporal-wise losses to learn scale-invariant representations. T-Rep \cite{fraikin2024t} further enhances temporal modeling by leveraging time-aware embeddings in the pretext task to capture fine-grained dependencies. SoftCLT \cite{lee2024soft} replaces the traditional hard contrastive objective with a soft contrastive loss that considers similarity scores across all negative pairs.

However, most existing methods neglect the inherent periodic structures present in many real-world time series. Moreover, approaches relying on dynamic time warping (DTW) to measure temporal similarity—such as SoftCLT—require a precomputed pairwise distance matrix, which becomes computationally prohibitive for long-term multivariate time series data.

\textbf{Latent state representation in time series.} \: Latent states—such as motion states in human activity recognition (HAR) or clinical states in healthcare—play a crucial role in characterizing the dynamics of time series data. Learning how these states evolve over time is essential for capturing long-term trajectories and predicting future trends.
To model such latent states, TNC \cite{tonekaboni2021unsupervised} introduces the notion of temporal neighborhoods, treating temporally adjacent windows as positive pairs and distant windows as negative pairs to encourage local consistency in the learned representations. Time2State \cite{wang2023time2state} proposes an unsupervised framework that applies a sliding window mechanism to extract distinguishable representations from temporally heterogeneous sequences. However, existing methods focus primarily on identifying latent states in isolation and neglect the similarity and transitions between them—which are critical for modeling temporal dependencies and state dynamics in real-world data. As a result, these approaches often yield coarse-grained representations that perform well for classification but generalize poorly to other downstream tasks such as forecasting or anomaly detection.

\section{Methods}
\label{headings}

In this section, we present PLanTS for learning both latent states and their dynamic transitions from multivariate time series in a self-supervised manner. An overview of the architecture is shown in Figure~\ref{fig2}b.
PLanTS comprises three main components: (1) a periodicity-aware multi-granularity patching module that decomposes raw time series into structured segments aligned with dominant periodic patterns; (2) two dedicated encoders—Latent State Encoder (LSE) and Dynamic Transition Encoder (DTE)—that extract complementary representations; and (3) a fusion step that combines both latent state and transition embeddings for downstream tasks.


\subsection{Problem definition}

Consider the multivariate time series input $X=\{x_1,x_2,...,x_N\}\in \mathbb{R}^{N\times L \times C}$, where $N$ denotes the number of samples, $L$ represents the number of timestamps and $C$ is the number of channels. The objective is to learn a non-linear embedding function $f_\theta$ to map each input sample $x_i$ into a latent representation $z_i\in \mathbb{R}^{L \times D}$, where $D$ is the embedding dimension. In PLanTS,  $f_\theta$ is composed of two sub-modules: the Latent State Encoder $f_L:\mathbb{R}^{L\times C} \rightarrow \mathbb{R}^{L\times D_l}
$, which captures latent states and is learned via a multi-granularity generalized contrastive loss. The Dynamic Transition Encoder $f_T:\mathbb{R}^{L\times C} \rightarrow \mathbb{R}^{L\times D_t}$, which models temporal transitions between latent states using a novel self-supervised pretext task. The final representation is computed as:
$z_i= \mathrm{concat}(f_L(x_i), f_T(x_i))$, where $D =D_l+D_t$.

\subsection{periodicity-aware multi-granularity patching mechanism}

\begin{figure}[t]
\centering
\includegraphics[width=1\textwidth]{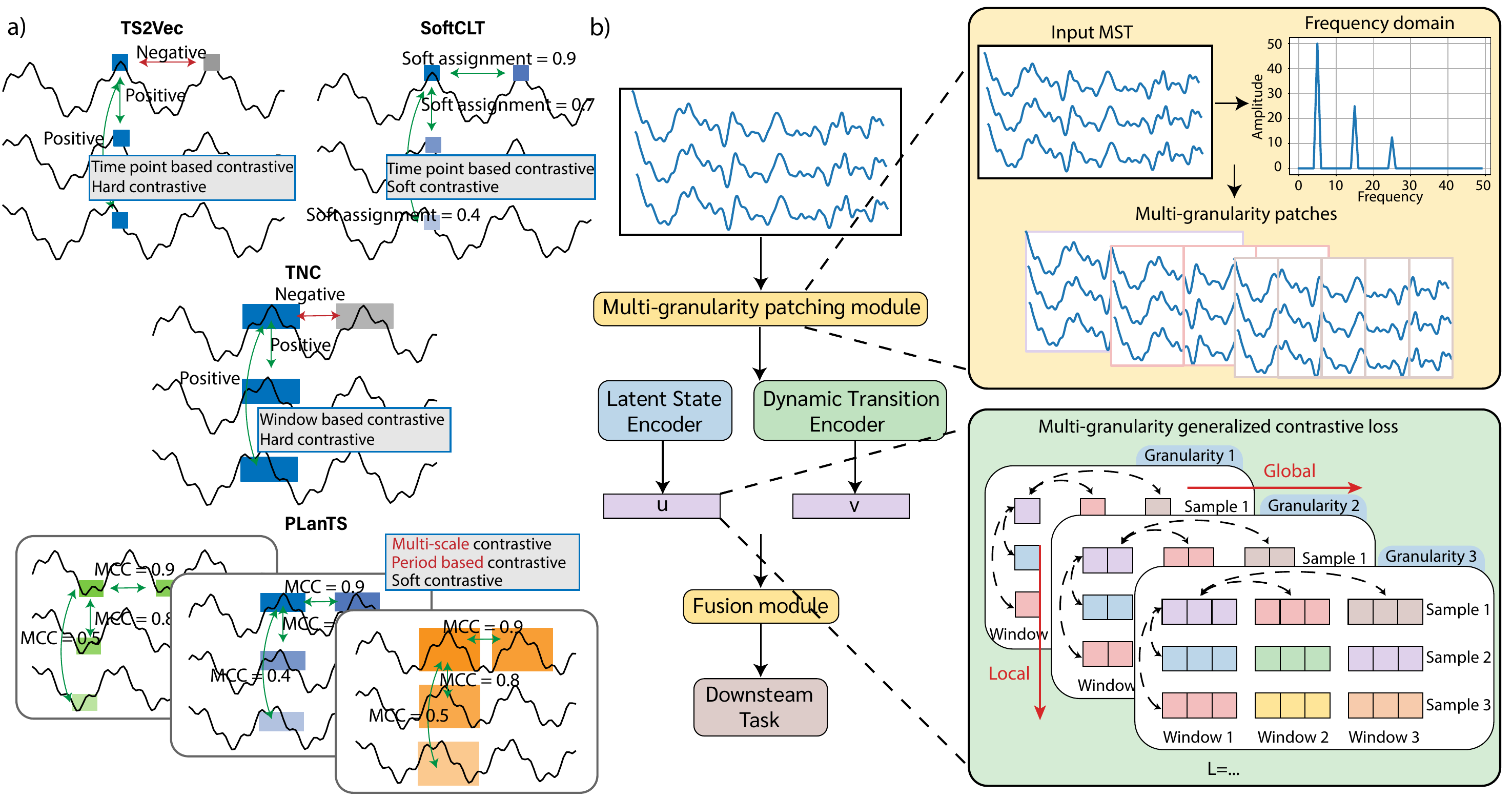}
\caption{
\textbf{Overview of the PLanTS framework.} a) Comparison with existing contrastive mechanisms for MTS. TS2Vec and SoftCLT utilize point-based contrastive learning, forming positive and negative pairs via contextual or soft assignment strategies. TNC applies a hard contrastive mechanism over fixed-size windows. In contrast, PLanTS incorporates periodic structure and introduces a multi-granularity, period-aware soft contrastive learning framework that operates on dynamic latent states. b) Overall PLanTS framework. 
}
\label{fig2}
\end{figure}

Multivariate time series often exhibit multiple latent states, each associated with distinct temporal patterns that unfold at irregular and variable time scales. Unlike words in a sentence, individual time points in a sequence lack semantic meaning, making it difficult for point-level contrastive learning methods \cite{yue2022ts2vec,lee2024soft} to capture meaningful latent states—let alone model transitions between latent states.
To capture latent states, TNC \cite{tonekaboni2021unsupervised} adopted a fixed-size window-based contrastive approach to model evolving statistical properties(Figure~\ref{fig2}a). However, choosing an appropriate window size is non-trivial: it is typically treated as a static hyperparameter, which cannot adapt to the heterogeneous and non-stationary nature of latent state dynamics across time.
To overcome this limitation, we propose a periodicity-aware multi-granularity patching mechanism that automatically selects meaningful window sizes based on the dominant periodic structures inferred from input time series.

Inspired by \cite{wu2022timesnet}, we employ the Fast Fourier Transform (FFT) to identify prominent periodic patterns and determine appropriate time scales for patching.

\begin{equation}
    F = \mathrm{Avg}(\mathrm{Amp}(\mathrm{FFT}(X))), \quad
    f_1, \dots, f_K = \arg\max_{\substack{f_* \in [1, \frac{L}{3}]}}^{\text{Top-}K}(F), \quad
    w_j = \lceil \frac{L}{f_j} \rceil
\end{equation}

Here, $\mathrm{FFT}(\cdot)$ denotes the Fourier transform applied along the temporal axis and $\mathrm{Amp}(\cdot)$ computes the corresponding amplitude spectrum. The vector $F \in \mathbb{R}^L$ represents the frequency-wise amplitude values averaged over all channels via $\mathrm{Avg}(.)$.
Note that the $j$-th value $F_j$ denotes the intensity of the $j$-th frequency periodic basis function, corresponding to a period length of $\lceil \frac{L}{f_j} \rceil$. To mitigate the influence of high-frequency noise, we retain the top-$K$ 
dominant frequency components $\{f_1,...,f_K\}$ based on their amplitudes values $\{F_1,...,F_K\}$. The associated period lengths $\{w_1,...w_K\}$ are computed and serve as window sizes for the multi-granularity patching module. 

Given an input multivariate time series sample $x_{i}\in \mathbb{R}^{L\times C}$ and the set of window sizes $\{w_1,...w_K\}$, we segment the input into non-overlapping patches for each granularity. For the $k$-th granularity, the series is divided into $M_k=\lceil \frac{L}{w_k}\rceil$ patches $x_p^k\in \mathbb{R}^{M_k\times w_k \times C} $. Zero-padding is applied if necessary to ensure divisibility. The resulting set of patches $x_p=\{x^{(1)},...,x^{(K)}\}$ is then fed into the Latent State Encoder (LSE) and Dynamic Transition Encoder (DTE) to extract latent state representations and dynamic transition representations.

\subsection{Latent state representations}

To effectively capture latent states from multivariate time series, it is crucial to model the semantic similarity between different states. However, the relationships among latent states are often continuous and hierarchical, rather than strictly binary. To address this challenge, we propose a multi-granularity generalized contrastive loss, which captures both instance-level and state-level similarities across multiple temporal resolutions.

\textbf{Periodic Feature Similarity.} To capture the similarities between time series segments,  SoftCLT \cite{lee2024soft} relies on a computationally expensive precomputed dynamic time warping (DTW) distance matrix, which is computationally prohibitive for long-term multivariate time series data. Inspired by \cite{yang2023simper}, we avoids explicit alignment by computing Maximum Cross-Correlation (MXCorr) between time series windows in the input space. MXCorr provides an efficient approximation of latent state similarity while preserving temporal structure.

Let $x,y \in \mathbb{R}^{w\times C}$ be two multivariate time series windows of length $w$ and $C$ channels, The MXCorr is defined as:

\begin{equation}
\mathrm{MXCorr}(x, y) = \frac{1}{C} \sum_{c=1}^{C} \max_{\tau \in [0, w-1]} \mathrm{CC}_c(x^{(c)}, y^{(c)}; \tau)
\end{equation}

where $\max_\tau \mathrm{CC}_c (\cdot) $ is the maximum normalized cross-correlation across all possible time lags $\tau$ for channel $c$.
 We implement an efficient batched version of MXCorr and provide a detailed computational comparison with SoftCLT in Appendix~\ref{apdx:comp_time}. 

\textbf{Local instance-wise contrastive learning.} 
We hypothesize that latent states encode identity-specific characteristics—i.e., even when two time series samples are in the same latent state, their representations should remain distinguishable due to individual variations. To capture this intrinsic difference, we design a local instance-wise contrastive loss that emphasizes distinctions among samples within the same time window.

Let $u_i\in \mathbb{R}^{M_k\times w_k \times D_l}$ denote the latent state embeddings of $i$-th time series sample $x^{(k)}_i \in \mathbb{R}^{M_k\times w_k \times C}$ at $k$-th granularity. PlanTS treats all other samples in the batch as negative views weighted by input-space feature similarity. For the $i$-th time series sample at $m$-th window, the local instance-wise contrastive loss  can be formulated as:

\begin{equation}
    l^{i,m}_{\text{local}} = -\sum_{j=1,j\neq i}^{B} \frac{\mathrm{exp}(s_{ij}^m)}{\sum_{j'=1,j'\neq i}^{B} \mathrm{exp}(s_{ij'}^m)} \mathrm{log} \frac{\mathrm{exp}(u_i^m \cdot u_j^m)}{\sum_{j'=1,j'\neq i}^{B} \mathrm{exp} (u_i^m \cdot u_{j'}^m)}
\end{equation}

Here, $s_{ij}^m=\mathrm{MXCorr}(x^{(k)}_{i,m},x^{(k)}_{j,m})$ denotes the input-space similarity between samples $i$ and sample $j$ at $m$-th window. This formulation generalizes the InfoNCE loss\cite{oord2018representation}, which uses a hard assignment by treating all negative pairs equally.  In contrast, our method introduces a soft weighting scheme based on the input-space similarity $s_{ij}$, encouraging alignment between representations that are more similar in the original space. We demonstrate in Appendix~\ref{apdx:derivation} that minimizing our weighted contrastive loss is equivalent to minimizing the KL divergence between the predicted softmax distribution and the similarity-based target distribution.

\textbf{Global state-wise contrastive learning.} 
While latent states often remain stable within short temporal windows, they typically evolve over longer time horizons. To model these evolutions and learn discriminative representations that reflect temporal progression, we introduce a global state-wise contrastive loss that captures the continuous relationships among latent states along the time axis.

Let $u_i^m\in \mathbb{R}^{ w_k \times D_l}$ denote the latent state representation of the $m$-th window from the $i$-th time series sample at granularity $k$.  Similar to the local instance-wise contrastive loss, PLanTS compares this window against all other windows from the same sample using similarity-weighted alignment based on input-space periodic features. The global contrastive loss for the $m$-th window is defined as:

\begin{equation}
    l^{i,m}_{\text{global}} = -\sum_{n=1,n\neq m}^{M_k} \frac{\mathrm{exp}(a_{mn}^i)}{\sum_{n'=1,n'\neq m}^{M_k} \mathrm{exp}(a_{mn'}^i)} \mathrm{log} \frac{\mathrm{exp}(u_i^m \cdot u_i^n)}{\sum_{n'=1,n'\neq m}^{M_k} \mathrm{exp} (u_i^m \cdot u_{i}^{n'})}
\end{equation}

Where $a_{mn}^i=\mathrm{MXCorr}(x^{(k)}_{i,m},x^{(k)}_{i,n})$ denotes the input-space similarity between windows $m$ and $n$ of sample $i$. The overall contrastive loss at $k$-th granularity is the joint of the local and global contrastive losses:

\begin{equation}
L_l^{(k)} = \frac{1}{N \cdot M_k} \sum_{i=1}^{N} \sum_{m=1}^{M_k} \left( \alpha \cdot l_{\text{local}}^{i,m} + (1-\alpha ) \cdot l_{\text{global}}^{i,m} \right)
\label{equ:alpha}
\end{equation}

where $\alpha$ is a hyperparameter controlling the contribution of each loss.


\subsection{Dynamic transition representations}
Beyond learning representations that distinguish between latent states, it is essential to model the dynamics of state transitions to effectively track and forecast temporal trajectories in multivariate time series. To this end, we introduce a novel next-transition prediction pretext task that encourages the model to encode predictive information into the dynamic transition representations.

\textbf{Next transition prediction.} In real-world scenarios, dynamic behavior often evolves in a state-dependent manner. For example, fluctuations in a patient's vital signs vary depending on their underlying clinical condition (i.e., latent state). To capture such conditional dynamics, we propose a next transition prediction task that learns to forecast future transitions by conditioning on both the current latent state and the current dynamic transition representation.

Given a time series sample $x^{(k)}_i \in \mathbb{R}^{M_k\times w_k \times C}$ at $k$-th granularity, The Dynamic Transition Encoder $f_T$ outputs dynamic transition embedding: $v_i=f_T(x^{(k)}_i)\in \mathbb{R}^{M_k\times w_k \times D_t}$. At each window $m$, we concatenate the latent state representation $u_i^m$ and dynamic transition representation $v_i^m$, and feed the result into a prediction head $G: \mathbb{R}^{D_l+D_t}\rightarrow \mathbb{R}^{D_t}$,implemented as a two-layer MLP with ReLU activations. The objective is to minimize the mean squared error (MSE) between the predicted next transition and the ground-truth transition at window $m+1$:

\begin{equation}
L_t^{(k)} = \frac{1}{N \cdot (M_k-1)} \sum_{i=1}^{N}\sum_{m=1}^{M_k - 1} \left| G\left(\mathrm{concat}(u_i^m, v_i^m)\right) - v_i^{m+1} \right|^2
\end{equation}

This pretext task encourages the model to encode transition dynamics that are sensitive to the current latent state, enabling temporally aware representation learning.

\textbf{Final Objective.}
The overall loss for PLanTS combines both latent state representation learning and dynamic transition modeling across all granularities:

\begin{equation}
    L = \frac{1}{K} \sum_{k=1}^{K}\left(\lambda L_l^{(k)} + (1-\lambda) L_{t}^{(k)} \right)
\label{equ:lambda}
\end{equation}
where $\lambda$ is a hyperparameter controlling the contribution of each loss.

\section{EXPERIMENTS}
\label{others}
We conduct extensive experiments to evaluate the effectiveness of PLanTS across a range of downstream tasks for multivariate time series (MTS): (1) multi-class classification, (2) multi-label classification, (3) forecasting, (4) anomaly detection (in Appendix~\ref{apdx:anomaly}). In addition, we perform ablation studies to assess the contribution of each core component in PLanTS. Finally, we analyze the temporal trajectories of the learned representations to better understand how latent state transitions are captured and encoded in the representation space. Detailed experimental setups, additional results, and further analysis are provided in the Appendix~\ref{apdx:details}.

\subsection{multi-class classification}
We evaluate the instance-level representations learned by PLanTS on 30 benchmark datasets from the UEA multivariate time series classification archive \cite{bagnall2018uea}, covering diverse domains such as healthcare, sensor systems, speech, and human activity recognition.
We compare PLanTS against 8 SOTA self-supervised learning baselines: DTW \cite{chen2013dtw}, TST \cite{zerveas2021transformer}, TS-TCC \cite{eldele2021time}, T-Loss \cite{franceschi2019unsupervised}, TNC \cite{tonekaboni2021unsupervised}, TS2Vec \cite{yue2022ts2vec}, T-Rep \cite{fraikin2024t}, and SoftCLT \cite{lee2024soft}. Following the evaluation protocol of TS2Vec, we train an SVM classifier with an RBF kernel on top of the learned representations to perform classification.

The evaluation results are summarized in Table~\ref{tab:uea_summary} and full results are shown in Appendix~\ref{subsec:full_results}. PLanTS achieves consistent and substantial improvements over all baselines. In particular, it improves the average classification accuracy by 2.3\% over TS2Vec, and achieves 2.0\% and 1.6\% gains over T-Rep and SoftCLT, respectively. Moreover, PLanTS obtains the best average rank and the second-highest number of first-place rankings, demonstrating strong performance in MTS classification task.

\begin{table}[t]
\centering
\begin{minipage}{0.48\textwidth}
\centering
\resizebox{\textwidth}{!}{
\begin{tabular}{lcccc}
\toprule
Method & Avg. Acc. & Avg. Rank & Ranks 1\textsuperscript{st} & Avg. Diff. (\%) \\
\midrule
DTW & 0.650 & 5.517 & 1 & 9.214 \\
TST & 0.617 & 6.300 & 1 & 10.863 \\
TS-TCC & 0.668 & 5.333 & 4 & 6.463 \\
T-Loss & 0.658 & 4.600 & 4 & 7.670 \\
TNC & 0.670 & 5.533 & 1 & 5.640 \\
TS2Vec & 0.704 & \underline{3.533} & 4 & 5.063 \\
T-Rep & 0.706 & 4.067 & \textbf{10} & 5.430 \\
SoftCLT & \underline{0.709} & 4.222 & 5 & 4.544 \\
\midrule
\rowcolor{gray!20}
PLanTS & \textbf{0.720} & \textbf{3.400} & \underline{8} & \textbf{--} \\
\bottomrule
\end{tabular}
}
\caption{Summary of classification results on the 30 UEA MTS archive.}
\label{tab:uea_summary}
\end{minipage}
\hfill
\begin{minipage}{0.5\textwidth}
\centering
\resizebox{\textwidth}{!}{
\begin{tabular}{llccc}
\toprule
Task & Method & Accuracy & F1 Score & AUROC \\
\midrule
\multirow{4}{*}{\centering Diagnostic} 
    & Ts2Vec         & 0.447 & \underline{0.594} & 0.825 \\
    & T-rep          & 0.440 & 0.558 & \underline{0.836} \\
    & SimCLR + DBPM  & \textbf{0.458} & 0.583 & 0.806 \\
    \rowcolor{gray!20}
    & PLanTS           & \textbf{0.458} & \textbf{0.601} & \textbf{0.852} \\
\midrule
\multirow{4}{*}{\centering Form} 
    & Ts2Vec         & \underline{0.366} & \underline{0.509} & \underline{0.768} \\
    & T-rep          & 0.311 & 0.482 & 0.744 \\
    & SimCLR + DBPM  & 0.349 & 0.480 & 0.752 \\
    \rowcolor{gray!20}
    & PLanTS           & \textbf{0.385} & \textbf{0.514} & \textbf{0.784} \\
\midrule
\multirow{4}{*}{\centering Rhythm} 
    & Ts2Vec         & 0.791 & 0.825 & 0.833 \\
    & T-rep          & \textbf{0.819} & \textbf{0.853} & 0.833 \\
    & SimCLR + DBPM  & 0.808 & 0.837 & \underline{0.838} \\
    \rowcolor{gray!20}
    & PLanTS           & \textbf{0.819} & \underline{0.852} & \textbf{0.863} \\
\bottomrule
\end{tabular}
}
\caption{Performance comparison on PTB-XL multi-label classification tasks}
\label{tab:ptbxl_results}
\end{minipage}
\end{table}


\subsection{multi-label classification}
Unlike multi-class classification, multi-label classification does not assume class exclusivity, where multiple conditions, behaviors, or events can occur simultaneously. Thus, it provides a more realistic and stringent evaluation of time series representations by requiring models to capture diverse, overlapping patterns. We evaluate PLanTS on PTB-XL\cite{wagner2020ptb}, the largest publicly available clinical ECG waveform dataset. Based on ECG annotation schema, there are three multi-label classification tasks: Diagnostic (44 classes), Form (19 classes), and Rhythm (12 classes). 

We formulate the evaluation protocol by training a One-vs-Rest SVM classifier with an RBF kernel on top of the learned representations. We compare PLanTS against three SOTA self-supervised learning methods: Ts2Vec\cite{yue2022ts2vec}, T-Rep \cite{fraikin2024t} and DBPM\cite{lan2024towards}, a recently proposed SSL approach specifically designed for multi-label tasks. We employ four evaluation metrics: accuracy, F1 score (micro-averaged), AUROC (macro-averaged) and per-class AUROC. Results are reported in Table~\ref{tab:ptbxl_results} and detailed in Appendix~\ref{subsec:full_results}. 

\begin{figure}[h]
\centering
\includegraphics[width=0.8\textwidth]{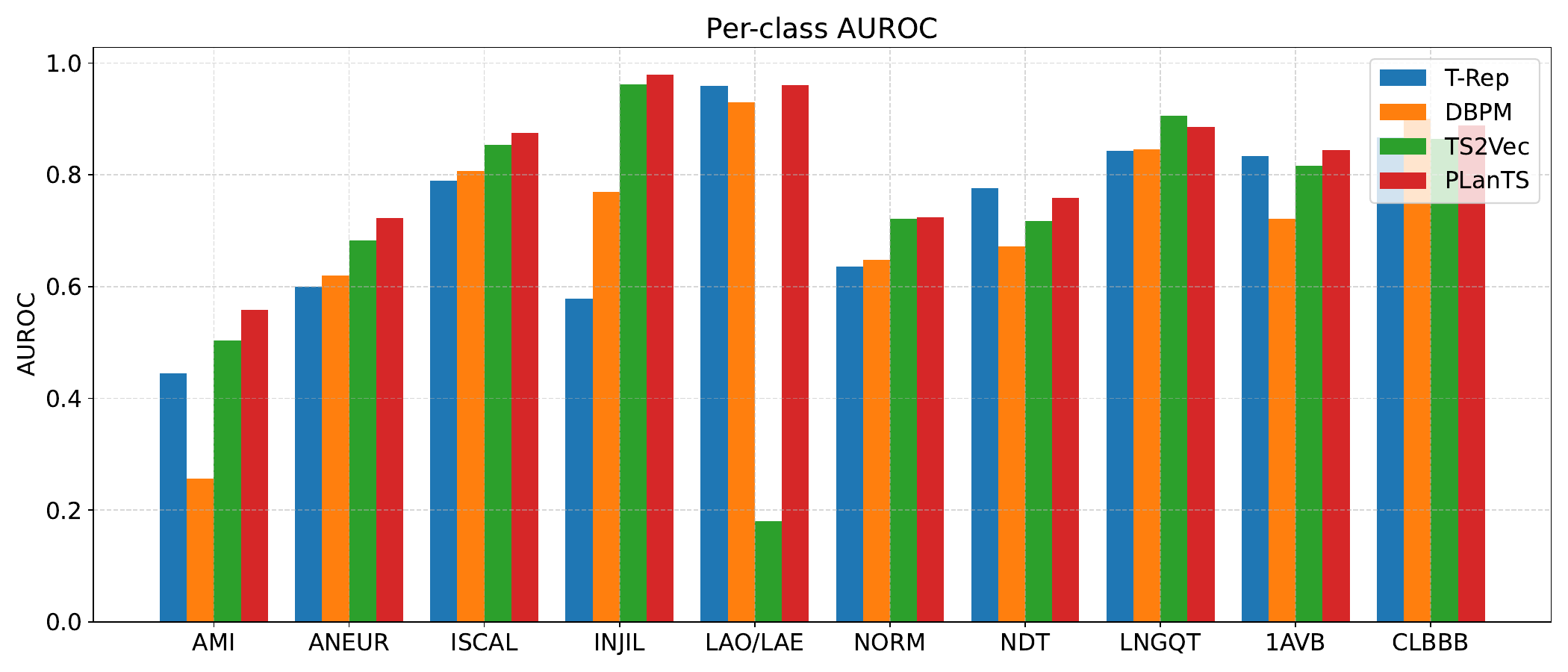}
\caption{Per-class AUROC comparison on 10 selected diagnostic classes from PTB-XL.}
\label{fig:barplot}
\end{figure}

PLanTS consistently achieves superior performance in terms of AUROC, improving from 0.836 to 0.852 in the Diagnostic task, from 0.768 to 0.784 in the Form task, and from 0.838 to 0.863 in the Rhythm task. For both Diagnostic and Form classification, PLanTS outperforms all baselines across all metrics. Specifically, it improves accuracy from 0.447 to 0.458 in the Diagnostic task and from 0.366 to 0.385 in the Form task. In the Rhythm task, PLanTS achieves the highest AUROC and closely approaching the top results in accuracy and F1 score. Figure~\ref{fig:barplot} shows the per-class AUROC for 10 selected diagnostic categories. While baseline methods suffer from performance drops in certain classes (e.g., DBPM on "AMI", T-Rep on "INJIL", and TS2Vec on "LAO/LAE"), PLanTS maintains consistently high AUROC across all categories. This highlights the robustness and reliability of PLanTS in capturing fine-grained clinical semantics in multivariate ECG data.

\subsection{FORECASTING}

We evaluate PLanTS on multivariate time series forecasting using four benchmark datasets from the ETT suite—ETTh1, ETTh2, ETTm1, and ETTm2—which include electricity load, oil temperature, and power consumption data collected over hourly and minute-level intervals \cite{zhou2021informer}. Following the protocol described in the [Appendix], we forecast multiple future horizons and report the averaged forecasting performance across all horizons. The proposed PLanTS model is compared to representative state-of-the-art methods such as TNC \cite{tonekaboni2021unsupervised}, TS2Vec \cite{yue2022ts2vec}, T-Rep \cite{fraikin2024t}, SoftCLT \cite{lee2024soft} and Informer\cite{zhou2021informer}. The average forecasting performances for each horizon, average rank and number of rank first over all datasets and prediction horizons are presented in Table 3 (full results are in Appendix~\ref{subsec:full_results}).

\begin{table}[t]
\centering
\resizebox{\textwidth}{!}{%
\begin{tabular}{lcccccccccccc}
\toprule
  & \multicolumn{2}{c}{PLanTS} & \multicolumn{2}{c}{SoftClt} & \multicolumn{2}{c}{T-rep} & \multicolumn{2}{c}{TS2Vec} & \multicolumn{2}{c}{Informer} & \multicolumn{2}{c}{TCN} \\
\cmidrule(lr){2-3} \cmidrule(lr){4-5} \cmidrule(lr){6-7} \cmidrule(lr){8-9} \cmidrule(lr){10-11} \cmidrule(lr){12-13}
 Dataset& MSE & MAE & MSE & MAE & MSE & MAE & MSE & MAE & MSE & MAE & MSE & MAE \\
\midrule
ETTh1 & \textbf{0.708} & \textbf{0.621} & 0.836 & 0.670 & \underline{0.763} & \underline{0.645} & 0.803 & 0.665 & 0.907 & 0.739 & 1.021 & 0.816 \\
ETTh2 & \underline{1.685} & \underline{0.967} & \textbf{1.494} & \textbf{0.925} & 1.818 & 1.034 & 1.802 & 1.022 & 2.371 & 1.199 & 2.574 & 1.265 \\
ETTm1 & \textbf{0.531} & \textbf{0.507} & 0.628 & 0.547 & \underline{0.584} & \underline{0.529} & 0.631 & 0.565 & 0.749 & 0.640 & 0.818 & 0.849 \\
ETTm2 & 0.885 & \underline{0.581} & \textbf{0.645} & \textbf{0.577} & \underline{0.783} & 0.603 & 0.784 & 0.607 & 1.173 & 0.702 & 3.635 & 1.891 \\
\midrule
Avg. Rank & \textbf{1.6} & \textbf{1.55} & 3.0 & 2.8 & 2.85 & 2.9 & 3.7 & 3.86 & 4.45 & 4.65 & 5.4 & 5.25 \\
Rank $1^{st}$ & \textbf{11} & \textbf{12} & \underline{5} & \underline{4} & 2 & 2 & 0 & 0 & 1 & 1 & 1 & 1 \\
\bottomrule
\end{tabular}
}  
\caption{Forecasting performance on the ETT benchmark.}
\label{tab:forecasting}
\end{table}

PLanTS achieves the best average performance overall, ranking first in 11 out of 16 settings (MSE) and 12 out of 16 settings (MAE). It consistently outperforms baseline methods on ETTh1 and ETTm1 in both MSE and MAE. On ETTh1 and ETTm1, PLanTS reduces the average MSE by 7.2\% and 9.1\%, and reduces MAE by 3.7\% and 4.2\%, respectively, compared to the strongest baseline (T-Rep). It also achieves competitive results on ETTh2. This demonstrates PLanTS's effectiveness in modeling fine-grained periodic and dynamic patterns for forecasting tasks. However, PLanTS does not perform as well on ETTm2 in terms of MSE. This may be due to the higher level of noise and abrupt fluctuations in ETTm2, which can degrade the quality of periodicity extraction and weaken the predictive strength of latent state transitions.


\subsection{Trajectory tracking}

To validate the ability of PLanTS to capture irregular latent state dynamics, we examine the temporal evolution of learned embeddings using the Human Activity Recognition (HAR) dataset from the UCI Machine Learning Repository\cite{anguita2013public}. UCI-HAR contains sensor data collected from 30 individuals wearing smartwatches while performing six activities: (1) walking, (2) walking upstairs, (3) walking downstairs, (4) sitting, (5) standing, and (6) laying.
Following the procedure in \cite{tonekaboni2021unsupervised}, we construct continuous activity trajectories for each subject by concatenating their activity sequences based on subject identifiers. This setup enables us to analyze transitions across multiple activity states in a realistic and temporally consistent manner.

\begin{figure}[h]
\centering
\includegraphics[width=0.7\textwidth]{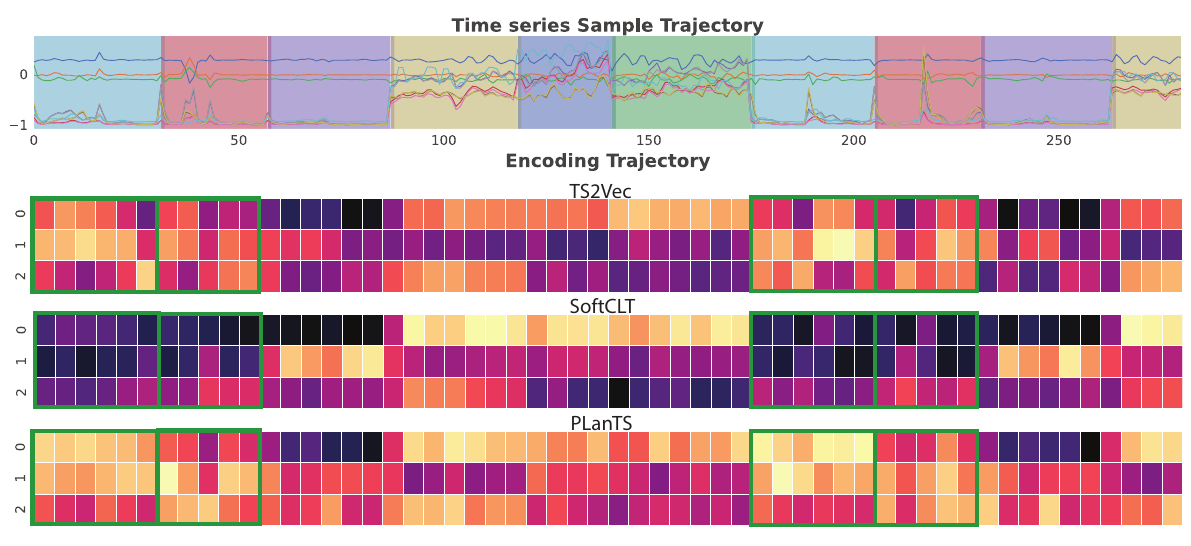}
\caption{Trajectory of a HAR signal encoding.}
\label{fig:heatmap}
\end{figure}

We visualize the top 3 principal components (PCA) of the learned embeddings and compare PLanTS with TS2Vec and SoftCLT. As shown in Figure~\ref{fig:heatmap}, PLanTS produces embeddings that exhibit sharper transitions and more distinct patterns across different activities. Notably, PLanTS better distinguishes between similar motion states such as \textit{sitting} and \textit{standing} (marked in red and cyan on the raw signal), which are difficult to separate in the original signals and baseline representations. As highlighted by green boxes in the embedding visualizations, these two states are clearly separable in PLanTS but remain indistinguishable in the representations of TS2Vec and SoftCLT.
Additional results are provided in Appendix~\ref{subsec:full_results}. The results indicate PLanTS's capability in modeling latent state transitions over time, which is a crucial property for post-hoc analysis and downstream applications, particularly in healthcare and human activity monitoring scenarios.

\subsection{Ablation study}

To assess the contribution of each component in PLanTS, we conduct comprehensive ablation studies on four forecasting datasets and four classification datasets. To better understand the architectural design of PLanTS, we compare its full version with the following variants:
\textbf{w/o multi-granularity patching:} removes the periodicity-aware multi-granularity patching mechanism and segments inputs into non-overlapping patches using a fixed window size of 50.
\textbf{w/o local contrastive:} disables the local instance-wise contrastive loss by setting $\alpha = 0$.
\textbf{w/o global contrastive:} disables the global state-wise contrastive loss by setting $\alpha = 1$.
\textbf{w/o NTP:} removes the next transition prediction pretext task by setting $\lambda = 1$.

Table~\ref{tab:ablation} summarizes the results. We observe that the multi-granularity patching mechanism is particularly beneficial for classification tasks. Removing it leads to consistent performance drops across all selected classification datasets. For instance, on the StandWalkJump dataset, classification accuracy drops by 50.07\%. In contrast, using a single fixed-size patching strategy slightly improves forecasting performance—e.g., MSE decreases by 2.88\%, 6.18\%, and 10.76\% on ETTh1, ETTh2, and ETTm1, respectively. This suggests that large periodicities in the ETT datasets might harm the model's fine-grained prediction ability.

Moreover, the contrastive losses and transition prediction task prove essential across both task types. In particular, removing the local contrastive loss leads to the largest degradation in classification performance, reducing accuracy by 60.02\% and 62.41\% on StandWalkJump and Handwriting, respectively. Eliminating the global contrastive loss causes a 7.24\% drop on Heartbeat. Removing the NTP objective decreases classification accuracy by 33.71\% on Handwriting and increases MSE by 6.79\% on ETTh2.

\begin{table}[t]
\centering
\resizebox{\textwidth}{!}{%
\begin{tabular}{lcccccccc}
\toprule
 & \multicolumn{4}{c}{Forecasting (↓ MSE)} & \multicolumn{4}{c}{Classification (↑ Accuracy)} \\
\cmidrule(lr){2-5} \cmidrule(lr){6-9}
Variant & ETTh1 & ETTh2 & ETTm1 & ETTm2 & StandWalkJump & Heartbeat & RacketSports & Handwriting \\
\midrule
PLanTS & 0.729 & 1.796 & 0.595 & 0.844 & \textbf{0.667} & \textbf{0.746} & \textbf{0.842} & \textbf{0.439} \\
w/o multi-granularity patching & \textbf{0.708} & \textbf{1.685} & \textbf{0.531} & 0.885 & 0.333 & 0.741 & 0.803 & 0.426 \\
w/o local contrastive & 0.795 & 1.916 & 0.571 & \textbf{0.826} & 0.200 & 0.746 & 0.796 & 0.165 \\
w/o global contrastive & 0.732 & 1.815 & 0.594 & 0.843 & 0.400 & 0.692 & 0.829 & 0.431 \\
w/o NTP & 0.735 & 1.918 & 0.571 & 0.849 & 0.333 & 0.737 & 0.829 & 0.291 \\
\bottomrule
\end{tabular}
}
\caption{Ablation study of PLanTS across forecasting and classification benchmarks.}
\label{tab:ablation}
\end{table}


\section{Conclusion}
In this paper, we propose PLanTS, a novel self-supervised learning framework to learn latent state representation for non-stationary Multivariate Time Series data. To model irregular latent states, we introduces a multi-granularity generalized contrastive loss guided by periodicity, which enable the model to preserve both instance-level and state-level similarities across multiple temporal resolutions. To further model the dynamics of state transitions, we design next-transition prediction pretext task that encourages the model to encode predictive information into the dynamic transition representations. We conducted extensive experiments across a wide range of downstream tasks—including multi-class and multi-label classification, forecasting, trajectory tracking and anomaly detection, which show significant improvement of performance. PlanTS has the ability to encode, track and predict the underlying latent state of MTS, which can be applicable in various downstream domains, especially in healthcare and human activity monitoring scenarios.



\bibliographystyle{plain}
\bibliography{references}

\newpage
\appendix

\section{Dataset Descriptions}
\textbf{Human Activity Recognition (HAR) dataset}

The UCI HAR dataset \cite{anguita2013public}is a widely used benchmark for human activity recognition tasks. It consists of sensor data collected from 30 subjects aged 19–48 while performing six activities of daily living: walking, walking upstairs, walking downstairs, sitting, standing, and laying. Each subject wore a Samsung Galaxy S II smartphone on their waist, which recorded tri-axial linear acceleration and angular velocity at a sampling rate of 50 Hz. The raw signals were segmented into fixed-width windows of 2.56 seconds (128 time steps) with a 50\% overlap. For each window, a set of 561 handcrafted time- and frequency-domain features was extracted. The dataset is split into training and test sets based on subject IDs. In our trajectory tracking experiment, we construct continuous activity trajectories for each subject by concatenating their activity sequences based on
subject identifiers. Details are shown in Table~\ref{tab:ptbxl}.

\textbf{PTB-XL ECG Database}

PTB-XL is a large-scale, publicly available electrocardiogram (ECG) dataset \cite{wagner2020ptb} published by the PhysioNet initiative. It contains 21,837 clinical 12-lead ECG records, each lasting 10 seconds and sampled at 500 Hz, from 18,885 unique patients. The dataset includes diagnostic annotations covering multiple labeling dimensions such as diagnostic, form, and rhythm classes, enabling both single- and multi-label classification tasks. Altogether, there are 71 distinct statements, comprised of 44 diagnostic, 12 rhythm, and 19 form statements, with 4 of these also serving as diagnostic ECG statements. Based on the ECG annotation method, there are three multi-label classification tasks: Diagnostic Classification (44 classes), Form Classification (19 classes), and Rhythm Classification (12 classes). We use data spliting rate 0.6,0.2,0.2 to split training, testing and validation sets and follow the data pre-processing steps from [ref]. Table~\ref{tab:ptbxl} provides a summarization of PTB-XL dataset.

\textbf{Yahoo dataset}

Yahoo dataset\cite{ren2019time} is a widely used benchmark for time-series anomaly detection, containing 367 synthetic and real-valued univariate time series grouped into four subsets (A1–A4), each labeled with point-wise anomalies. For fair comparison, we follow the same evaluation strategy as \cite{yue2022ts2vec}. The anomalies detected within a certain delay (7 steps for minutely data and 3 steps for hourly data) are considered correct. Additionally, during preprocessing, the raw time series is differenced $d$ times to mitigate non-stationary drift, where $d$ is the number of unit roots estimated using the Augmented Dickey-Fuller (ADF) test.

\begin{table}[t]
\centering
\label{tab:ptbxl}
\resizebox{0.6\textwidth}{!}{
\begin{tabular}{lcccccc}
\toprule
Dataset & Train & Val & Test & Channels & Length & Categories \\
\midrule
PTB-XL Diagnostic & 13688 & 3422 & 4278 & 12 & 1000 & 44 \\
PTB-XL Form & 5745 & 1437 & 1796 & 12 & 1000 & 19 \\
PTB-XL Rhythm & 13459 & 3365 & 4206 & 12 & 1000 & 12 \\
\midrule
HAR & 21 & -- & 9 & 561 & 281,288 & 6 \\
\bottomrule
\end{tabular}
}
\end{table}

\section{Implementation details}
\label{apdx:details}
The models are implemented in Python 3.12.11, using PyTorch 2.3.0 for deep learning and scikit-learn for SVMs, linear regressions, and data pre-processing. We employ the Adam optimizer in all experiments. Training is conducted on AWS g5 xlarge and g5 2xlarge instances, each equipped with NVIDIA A10G GPUs, using CUDA 11.6.

The hyperparameter configurations used in our experiments are summarized in  Table~\ref{tab:hypermeter}. There are five hyperparameters used in PLanTS: $\alpha$, $\lambda$, $K$, window size, learning rate ($lr$), and batch size ($bs$). Here, $\alpha$ and $\lambda$ control the relative contributions of the local contrastive, global contrastive, and next-transition prediction losses; we report them as pairs. $K$ denotes the number of dominant periodicities used in the period-aware multi-granularity patching strategy. When this mechanism is not applied, we instead report the fixed window size used. $lr$ represents learning rate and $bs$ denotes batch size. For $(\alpha, \lambda)$, we select from $\{(0.5, 0.5), (0.9, 1)\}$ depending on the task. We apply the period-aware multi-granularity patching mechanism in the \textit{Classification} and \textit{Trajectory Tracking} tasks, setting $K = 3$. For \textit{Multi-label Classification} and \textit{Forecasting}, we replace $K$ with fixed window sizes: [20, 30] for multi-label classification and 50 for forecasting. The learning rate is fixed at $0.001$ for all tasks except \textit{Classification}, where we sweep from $0.0001$ to $0.001$ to ensure convergence across all 30 UEA datasets. The batch size is set to $128$ for all experiments.  

\begin{table}[t]
\centering
\label{tab:hypermeter}
\resizebox{1\textwidth}{!}{
\begin{tabular}{|l|l|l|l|l|l|}
\hline
    Hyperparameter & Classification & Trajectory tracking &Anomaly detection& Multi-label classification& Forecasting \\
\hline
$(\alpha,\lambda)$ & (0.5,0.5),(0.9,1) & (0.5,0.5) & \multicolumn{2}{c|}{(0.9,1)} & (0.5,0.5) \\
\hline
$K$ & \multicolumn{3}{c|}{3} & window size=[20,30] & window size=50 \\
\hline
$lr$ & 0.0001-0.001 & \multicolumn{3}{c}{0.001} &  \\
\hline
$bs$ & \multicolumn{5}{c|}{128} \\
\hline
\end{tabular}
}
\caption{Hyperparameter settings for various tasks.}
\end{table}

\section{Hyper-parameter sensitivity}
\begin{figure}[h]
\centering
\includegraphics[width=1\textwidth]{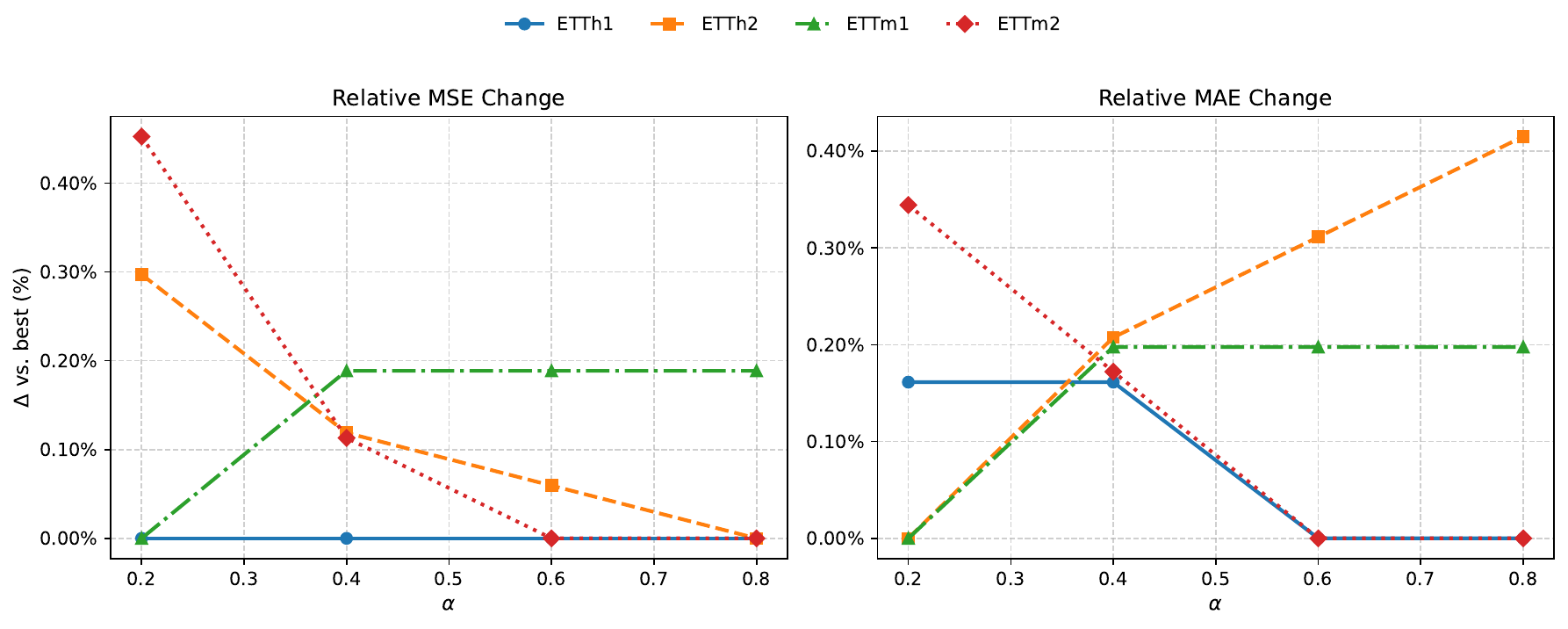}
\caption{Sensitivity analysis of hyper-parameters $\alpha$ in forecasting task.}
\label{fig:sensitivity1}
\end{figure}

\begin{figure}[h]
\centering
\includegraphics[width=1\textwidth]{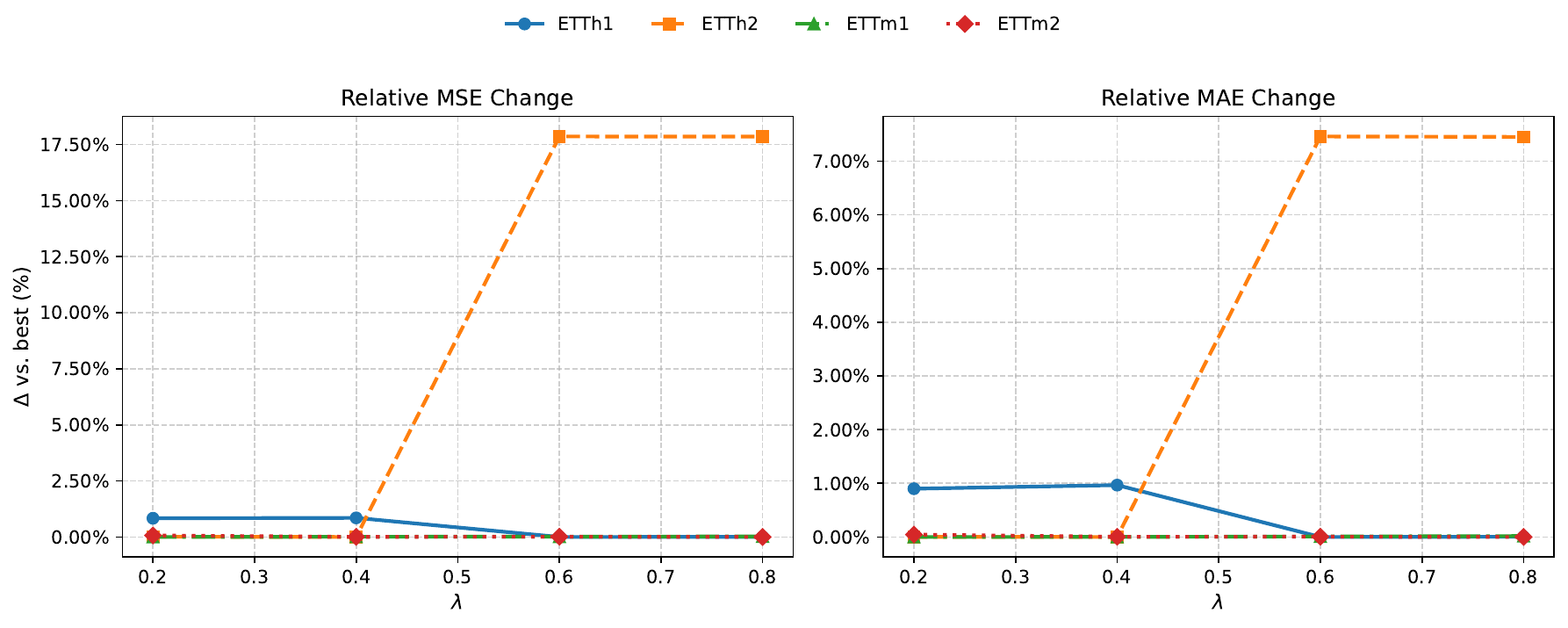}
\caption{Sensitivity analysis of hyper-parameters $\lambda$ in forecasting task.}
\label{fig:sensitivity2}
\end{figure}

We evaluate the sensitivity of PLanTS to the hyperparameters $\alpha$ and $\lambda$ (introduced in Equations~\ref{equ:alpha} and~\ref{equ:lambda}), which control the relative weights of the loss terms. Figures~\ref{fig:sensitivity1} and~\ref{fig:sensitivity2} report the relative percentage change in MSE and MAE with respect to the best results across four forecasting datasets.  
Overall, PLanTS exhibits stable performance under a wide range of hyperparameter values, demonstrating the robustness of the framework. We also observe that $\lambda$, which balances the latent state representation loss against the dynamic transition loss, has a stronger influence on performance—particularly on ETTh2—suggesting that accurately modeling transition dynamics is critical for datasets with more complex temporal dependencies.

\section{Derivation of weighted contrastive loss}
\label{apdx:derivation}

In this section, we aim at demonstrating that minimizing our weighted contrastive loss is equivalent to minimizing the KL divergence between the predicted softmax distribution and the similarity-based target distribution.
We define the predicted softmax distribution as $Q$, and define the input-space similarity distribution measured by Maximum Cross-Correlation as $P$, where: 

\begin{equation}
    \mathrm{q}(i,j)=\mathrm{Q}_{ij}=\frac{\mathrm{exp}(u_i^m \cdot u_j^m)}{\sum_{j'=1,j'\neq i}^{B} \mathrm{exp} (u_i^m \cdot u_{j'}^m)}
\end{equation}

\begin{equation}
    \mathrm{p}(i,j)=\mathrm{P}_{ij}=\frac{\mathrm{exp}(s_{ij}^m)}{\sum_{j'=1,j'\neq i}^{B} \mathrm{exp}(s_{ij'}^m)}
\end{equation}

Then the local instance-wise contrastive loss can be formulated as: 
\begin{align*}
     l^{i,m}_{\text{local}} &= -\sum_{j=1,j\neq i}^{B} \mathrm{p}(i,j)\:  \mathrm{log}\:  \mathrm{q}(i,j)
     \\ &= \sum_{j=1,j\neq i}^{B} (\mathrm{p}(i,j)\:  \mathrm{log}\:  \mathrm{p}(i,j)-\mathrm{p}(i,j)\:  \mathrm{log}\:  \mathrm{q}(i,j))-\sum_{j=1,j\neq i}^{B}\mathrm{p}(i,j)\:  \mathrm{log}\:  \mathrm{p}(i,j)
     \\ &=\sum_{j=1,j\neq i}^{B} \mathrm{p}(i,j)\: \mathrm{log}\: \frac{\mathrm{p}(i,j)}{\mathrm{q}(i,j)}-\sum_{j=1,j\neq i}^{B}\mathrm{p}(i,j)\:  \mathrm{log}\:  \mathrm{p}(i,j)
     \\ &=\mathrm{KL} (Q||P)+\mathrm{constant}
\end{align*}

\section{Computational comparison}
\label{apdx:comp_time}

To show the computation efficiency of PLanTS, we compare the running time of our method with one hard contrastive learning method Ts2Vec and one weighted contrastive learning method SpftCLT. All experiments are conducted on simulated data under controlled settings. For fairness, we adopt the TS2Vec backbone architecture, set the batch size to 128 for all methods, and use a single-granularity strategy for PLanTS in this comparison. To better investgate the effect of sequence length $L$, number of samples $N$ and number of channels $C$ to running time, we keep two other variables fixed (e.g., 5,000 samples, 3 channels), vary only one variable (e.g. sequence length). Runtime is decomposed into \emph{precomputation time}, \emph{training time}, and \emph{total runtime} (sum of both). 

Tables~\ref{tab:runtime_combined} summarize the results. For all settings, TS2Vec achieves the lowest overall runtime due to its hard contrastive strategy. When comparing PLanTS to SoftCLT, we observe a consistent advantage in total runtime despite PLanTS operating fully end-to-end without any precomputation phase. For example, at $L=256$, SoftCLT requires over $715$ seconds in total—driven largely by an expensive DTW-based precomputation step ($380.78 \pm 2.52$ seconds)—whereas PLanTS completes training in $110.93 \pm 0.61$ seconds. This advantage is maintained for longer sequences: at $L=1024$, SoftCLT takes $650.23 \pm 2.57$ seconds, while PLanTS requires only $234.07 \pm 0.37$ seconds. Similar trends are observed when scaling the number of samples or channels, confirming the scalability and computational efficiency of PLanTS.

\begin{table}[t]
\centering
\label{tab:runtime_combined}
\resizebox{0.8\textwidth}{!}{
\begin{tabular}{cclccc}
\toprule
Varied & Value & Method & Precomp & Train & Total \\
\midrule
\multirow{9}{*}{$L$} 
 & 256  & TS2Vec  & -- & $26.34_{\pm1.32}$ & $26.34_{\pm1.32}$ \\
 &      & PLanTS  & -- & $110.93_{\pm0.61}$ & $110.93_{\pm0.61}$ \\
 &      & SoftCLT & $380.78_{\pm2.52}$ & $335.17_{\pm4.78}$ & $715.01_{\pm8.88}$ \\
\cmidrule{2-6}
 & 512  & TS2Vec  & -- & $57.55_{\pm0.81}$ & $57.55_{\pm0.81}$ \\
 &      & PLanTS  & -- & $166.26_{\pm0.27}$ & $166.26_{\pm0.27}$ \\
 &      & SoftCLT & $24.90_{\pm0.09}$ & $447.17_{\pm4.19}$ & $472.08_{\pm4.11}$ \\
\cmidrule{2-6}
 & 1024 & TS2Vec  & -- & $98.03_{\pm3.02}$ & $98.03_{\pm3.02}$ \\
 &      & PLanTS  & -- & $234.07_{\pm0.37}$ & $234.07_{\pm0.37}$ \\
 &      & SoftCLT & $25.99_{\pm0.17}$ & $624.23_{\pm2.40}$ & $650.23_{\pm2.57}$ \\
\midrule
\multirow{9}{*}{$N$} 
 & 100  & TS2Vec  & -- & $17.62_{\pm2.80}$ & $17.62_{\pm2.80}$ \\
 &      & PLanTS  & -- & $45.31_{\pm0.85}$ & $45.31_{\pm0.85}$ \\
 &      & SoftCLT & $1.21_{\pm0.35}$ & $91.28_{\pm7.57}$ & $92.48_{\pm7.92}$ \\
\cmidrule{2-6}
 & 500  & TS2Vec  & -- & $57.55_{\pm0.81}$ & $57.55_{\pm0.81}$ \\
 &      & PLanTS  & -- & $166.26_{\pm0.27}$ & $166.26_{\pm0.27}$ \\
 &      & SoftCLT & $24.90_{\pm0.09}$ & $447.17_{\pm4.19}$ & $472.08_{\pm4.11}$ \\
\cmidrule{2-6}
 & 1000 & TS2Vec  & -- & $128.12_{\pm2.39}$ & $128.12_{\pm2.39}$ \\ 
 &      & PLanTS  & -- & $387.98_{\pm0.13}$ & $387.98_{\pm0.13}$ \\
 &      & SoftCLT & $99.31_{\pm0.43}$ & $899.07_{\pm13.02}$ & $998.39_{\pm12.94}$ \\
\midrule
\multirow{9}{*}{$C$} 
 & 3    & TS2Vec  & -- & $57.55_{\pm0.81}$ & $57.55_{\pm0.81}$ \\
 &      & PLanTS  & -- & $166.26_{\pm0.27}$ & $166.26_{\pm0.27}$ \\
 &      & SoftCLT & $24.90_{\pm0.09}$ & $447.17_{\pm4.19}$ & $472.08_{\pm4.11}$ \\
\cmidrule{2-6}
 & 10   & TS2Vec  & -- & $57.44_{\pm4.68}$ & $57.44_{\pm4.68}$ \\
 &      & PLanTS  & -- & $170.59_{\pm0.53}$ & $170.59_{\pm0.53}$ \\
 &      & SoftCLT & $40.95_{\pm0.39}$ & $448.41_{\pm7.56}$ & $489.36_{\pm7.93}$ \\
\cmidrule{2-6}
 & 20   & TS2Vec  & -- & $58.24_{\pm5.07}$ & $58.24_{\pm5.07}$ \\ 
 &      & PLanTS  & -- & $178.64_{\pm0.58}$ & $178.64_{\pm0.58}$ \\
 &      & SoftCLT & $61.11_{\pm0.55}$ & $456.37_{\pm10.39}$ & $517.48_{\pm10.95}$ \\
\bottomrule
\end{tabular}
}
\caption{End-to-end runtime comparison on simulated data under varying sequence length $L$, number of samples $N$, and number of channels $C$. All times are in \textbf{seconds}. Precomp: precomputation time. Train: training time. Total: sum of both.}
\end{table}

\section{Anomaly detection task}
\label{apdx:anomaly}

We preform point-based anomaly detection experiment on Yahoo dataset\cite{ren2019time}. We follow the evaluation protocol of \cite{yue2022ts2vec}. Given time series slice $x_1, x_2 ,...,x_t$, the target is to determine whether the last time point $x_t$ is an anomaly. The anomaly score is computed as the $L_1$ distance between representations with masked and unmasked input. We evaluate PLanTS under two experiment setting: normal setting and cold-start setting, and compare results against 11 baseline methods. For normal setting, we consider SPOT, DSPOT ,DONUT and SR. For cold-start setting, we compare wtih FFT, Twitter-AD, Luminol and SR. We also use SSL methods:TS2Vec, T-Rep and SoftCLT as baseline methods for both settings. The results are reported in Table~\ref{tab:anomaly}. From the results, PLanTS outperforms all the baseline methods in terms of F1 score. Remarkably, PLanTS improves F1 score approximately 2\% with respect to SoftCLT and TS2Vec.

\begin{table}[t]
\centering
\label{tab:anomaly}
\resizebox{0.6\textwidth}{!}{%
\begin{tabular}{lccclccc}
\toprule
 & \multicolumn{3}{c}{Yahoo Normal} & & \multicolumn{3}{c}{Yahoo Cold Start} \\
\cmidrule(lr){2-4} \cmidrule(lr){6-8}
Method & F1&	Prec&	Rec&	Method&	F1&	Prec&	Rec \\
\midrule
SPOT&   	33.8&	26.9&	45.4& FFT&	29.1&	20.2&	51.7\\
DSPOT&	31.6&	24.1&	45.8&	Twitter-AD&	24.5&	16.6&	46.2\\
DONUT&	2.6&	1.3&	82.5&	Luminol&	38.8&	25.4&	81.8\\
SR&	5.63&	45.1&	74.7&	SR&	52.9&	40.4&	76.5\\
TS2Vec&	74.5&	72.9&	76.2&	TS2Vec&	72.6&	69.2&	76.3\\
T-Rep&	75.7&	81.0&	74.5&	T-Rep&	76.3&	79.4&	73.4\\
SoftCLT&	74.2&	72.2&	76.5&	SoftCLT&	76.2&	75.3&	77.3\\
PLanTS&	\textbf{77.3}&	84.1&	71.5&	PLanTS&	\textbf{77.4}&	83.7&	72.0\\
\bottomrule
\end{tabular}
}
\caption{Time series anomaly detection results.}
\end{table}

\section{Full results}

\begin{figure}[h]
\centering
\includegraphics[width=1\textwidth]{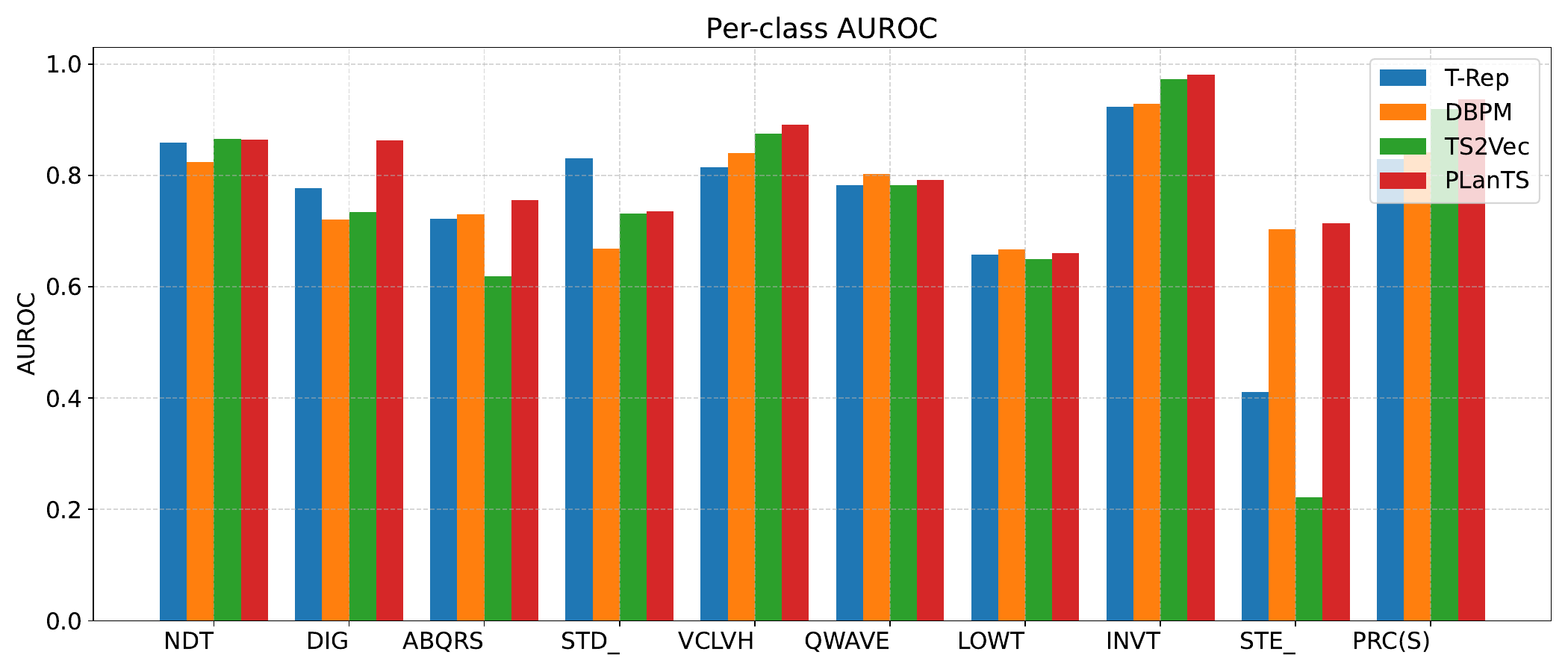}
\caption{Per-class AUROC comparison on 10 selected form classes from PTB-XL.}
\label{fig:sample}
\end{figure}

\begin{figure}[h]
\centering
\includegraphics[width=1\textwidth]{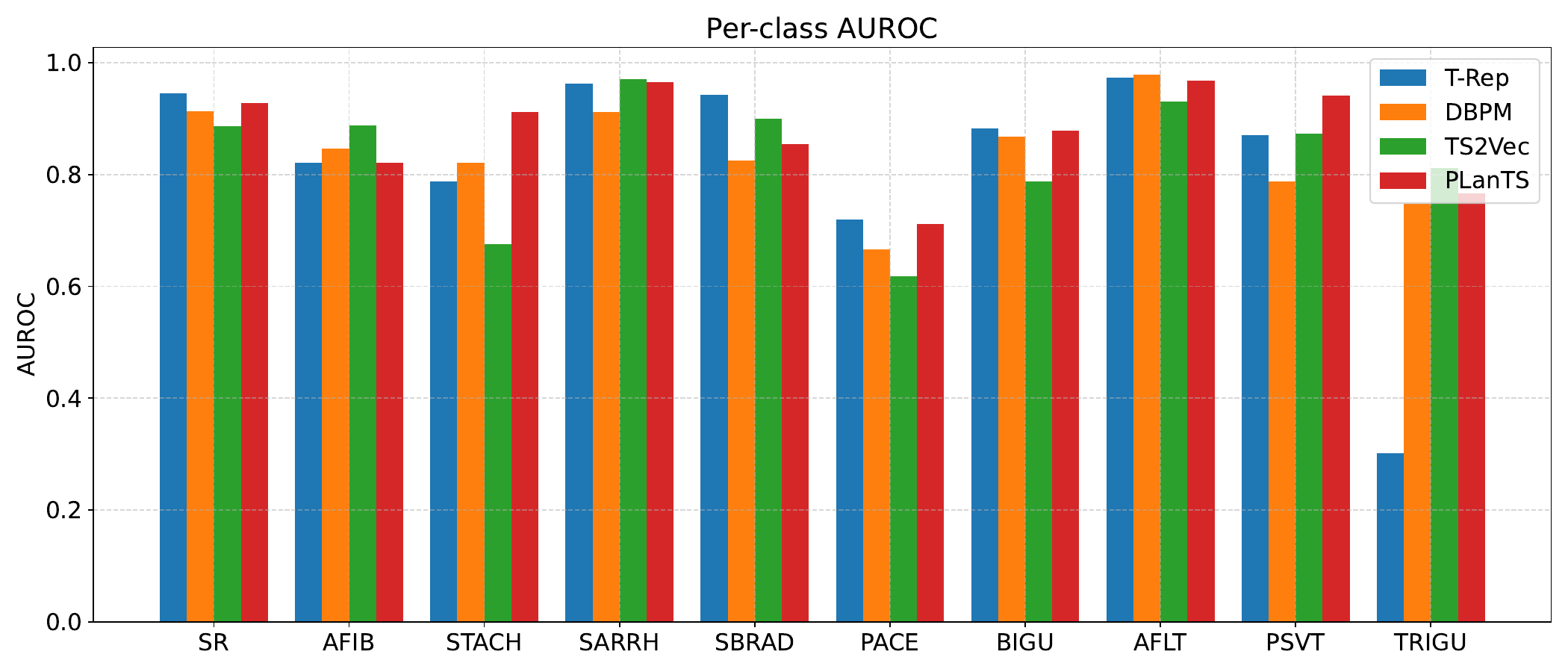}
\caption{Per-class AUROC comparison on 10 selected rhythm classes from PTB-XL.}
\label{fig:sample2}
\end{figure}

\begin{figure}[h]
\centering
\includegraphics[width=1\textwidth]{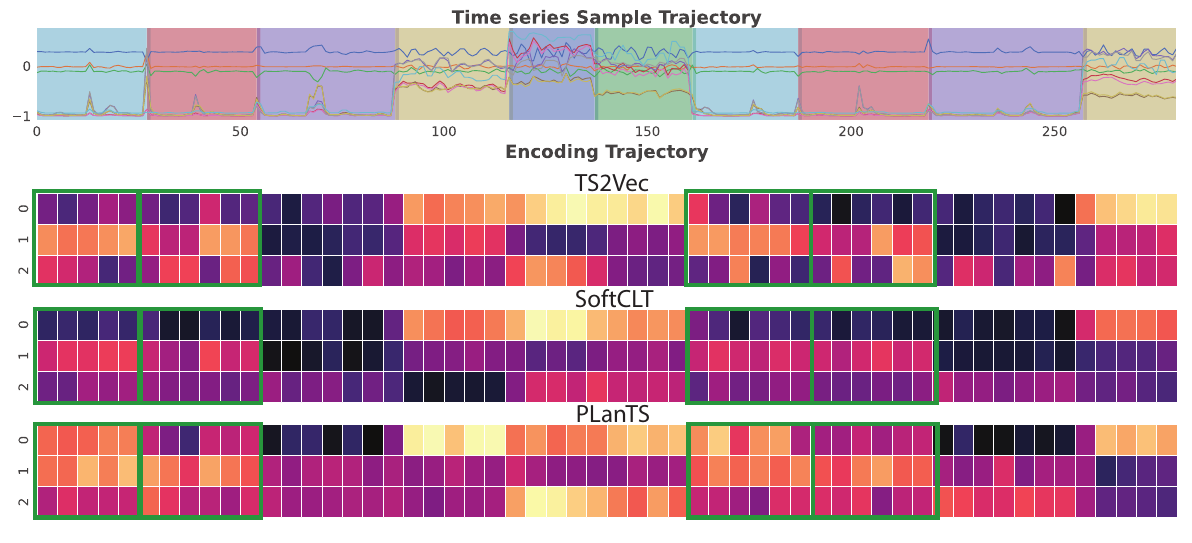}
\caption{Trajectory of another HAR signal encoding.}
\label{fig:heatmap2}
\end{figure}

The full results of MTS classification task on 30 UEA datasets are shown in Table~\ref{tab:uea_cla_full}. The full results for the forecasting task on the 4 ETT datasets are presented in Table~\ref{tab:forecasting}. Figure~\ref{fig:sample} and Figure~\ref{fig:sample2} shows the per-class AUROC for 10 selected form categories and 10 selected rhythm categories, respectively. For trajectory tracking task, Figure~\ref{fig:heatmap2} shows another example of comparison among top 3 principal components (PCA) of the learned embeddings of PLanTS, TS2Vec and SoftCLT.



\label{subsec:full_results}

\begin{table}[t]
\centering
\label{tab:uea_cla_full}
\resizebox{\textwidth}{!}{%
\begin{tabular}{lccccccccc}
\toprule
Dataset & PLanTS & softclt & T-Rep & TS2vec & T-Loss & TNC & TS-TCC & TST & DTW \\
\midrule
ArticularyWordRecognition & 0.973 &    0.990 &  0.968 &   0.987 &   0.943 & 0.973 &   0.953 & 0.977 & 0.987 \\
       AtrialFibrillation & 0.267 &    0.200 &  0.354 &   0.200 &   0.133 & 0.133 &   0.267 & 0.067 & 0.200 \\
             BasicMotions & 1.000 &    0.975 &  1.000 &   0.975 &   1.000 & 0.975 &   1.000 & 0.975 & 0.975 \\
    CharacterTrajectories & 0.983 &    0.992 &  0.989 &   0.995 &   0.993 & 0.967 &   0.985 & 0.975 & 0.989 \\
                  Cricket & 1.000 &    0.972 &  0.958 &   0.972 &   0.972 & 0.958 &   0.917 & 1.000 & 1.000 \\
            DuckDuckGeese & 0.560 &    0.360 &  0.457 &   0.680 &   0.650 & 0.460 &   0.380 & 0.620 & 0.600 \\
               EigenWorms & 0.809 &      -- &  0.884 &   0.847 &   0.840 & 0.840 &   0.779 & 0.748 & 0.618 \\
                 Epilepsy & 0.971 &    0.942 &  0.970 &   0.964 &   0.971 & 0.957 &   0.957 & 0.949 & 0.964 \\
                    ERing & 0.852 &    0.941 &  0.943 &   0.874 &   0.133 & 0.852 &   0.904 & 0.874 & 0.133 \\
     EthanolConcentration & 0.274 &    0.278 &  0.333 &   0.308 &   0.205 & 0.297 &   0.285 & 0.262 & 0.323 \\
            FaceDetection & 0.550 &    0.493 &  0.581 &   0.501 &   0.513 & 0.536 &   0.544 & 0.534 & 0.529 \\
          FingerMovements & 0.580 &    0.580 &  0.495 &   0.480 &   0.580 & 0.470 &   0.460 & 0.560 & 0.530 \\
    HandMovementDirection & 0.446 &    0.392 &  0.536 &   0.338 &   0.351 & 0.324 &   0.243 & 0.243 & 0.231 \\
              Handwriting & 0.439 &    0.467 &  0.414 &   0.515 &   0.451 & 0.249 &   0.498 & 0.225 & 0.286 \\
                Heartbeat & 0.746 &    0.722 &  0.725 &   0.683 &   0.741 & 0.746 &   0.751 & 0.746 & 0.717 \\
           JapaneseVowels & 0.976 &    0.978 &  0.962 &   0.984 &   0.989 & 0.978 &   0.930 & 0.978 & 0.949 \\
                   Libras & 0.861 &    0.889 &  0.829 &   0.867 &   0.883 & 0.817 &   0.822 & 0.656 & 0.870 \\
                     LSST & 0.598 &    0.534 &  0.526 &   0.537 &   0.509 & 0.595 &   0.474 & 0.408 & 0.551 \\
             MotorImagery & 0.570 &      -- &  0.495 &   0.510 &   0.580 & 0.500 &   0.610 & 0.500 & 0.500 \\
                   NATOPS & 0.917 &    0.944 &  0.804 &   0.928 &   0.917 & 0.911 &   0.822 & 0.850 & 0.883 \\
                  PEMS-SF & 0.803 &    0.723 &  0.800 &   0.682 &   0.675 & 0.699 &   0.734 & 0.740 & 0.711 \\
                PenDigits & 0.986 &    0.987 &  0.971 &   0.989 &   0.981 & 0.979 &   0.974 & 0.560 & 0.977 \\
           PhonemeSpectra & 0.247 &    0.223 &  0.232 &   0.233 &   0.222 & 0.207 &   0.252 & 0.085 & 0.151 \\
             RacketSports & 0.842 &    0.855 &  0.883 &   0.855 &   0.855 & 0.776 &   0.816 & 0.809 & 0.803 \\
       SelfRegulationSCP1 & 0.901 &    0.799 &  0.819 &   0.812 &   0.843 & 0.799 &   0.823 & 0.754 & 0.775 \\
       SelfRegulationSCP2 & 0.544 &    0.500 &  0.591 &   0.578 &   0.539 & 0.550 &   0.533 & 0.550 & 0.539 \\
       SpokenArabicDigits & 0.951 &    0.949 &  0.994 &   0.988 &   0.905 & 0.934 &   0.970 & 0.923 & 0.963 \\
            StandWalkJump & 0.667 &    0.533 &  0.441 &   0.467 &   0.332 & 0.400 &   0.333 & 0.267 & 0.200 \\
      UWaveGestureLibrary & 0.850 &    0.925 &  0.885 &   0.906 &   0.875 & 0.759 &   0.753 & 0.575 & 0.903 \\
           InsectWingbeat & 0.423 &      -- &  0.328 &   0.466 &   0.156 & 0.469 &   0.264 & 0.105 &   -- \\
\midrule
Avg.Acc. & \textbf{0.720} &      0.709 &  0.706 &   0.704 &   0.670 & 0.658 &   0.668 & 0.617 &   0.650 \\
Avg. Rank & \textbf{3.4} & 4.2 & 4.1& 3.5& 5.5& 4.6& 5.3& 6.3& 5.5 \\

Ranks 1\textsuperscript{st} & 8 & 5 & \textbf{10} & 4 & 1& 4& 4& 1 &1\\
\bottomrule
\end{tabular}%
}
\caption{Full classification results on 30 UEA datasets.}
\end{table}

\begin{table}[t]
\centering
\label{tab:forecasting}
\resizebox{\textwidth}{!}{%
\begin{tabular}{llcccccccccccc}
\toprule
 &  & \multicolumn{2}{c}{PLanTS} & \multicolumn{2}{c}{SoftClt} & \multicolumn{2}{c}{T-rep} & \multicolumn{2}{c}{TS2Vec} & \multicolumn{2}{c}{Informer} & \multicolumn{2}{c}{TCN} \\
 \cline{3-14}
 Dataset&H  & MSE & MAE & MSE & MAE & MSE & MAE & MSE & MAE & MSE & MAE & MSE & MAE \\
\midrule
\multirow{5}{*}{\centering ETTh1} 
    & 24 & 0.518 & 0.508 & 0.630 & 0.550 & \textbf{0.511} & \textbf{0.496} & 0.575 & 0.529 & 0.577 & 0.549 & 0.767 & 0.612 \\
    & 480 & 0.547 & 0.529 & 0.670 & 0.579 & \textbf{0.546} & \textbf{0.524} & 0.608 & 0.553 & 0.685 & 0.625 & 0.713 & 0.617 \\
    & 168 & \textbf{0.676} & \textbf{0.607} & 0.814 & 0.664 & 0.759 & 0.649 & 0.782 & 0.659 & 0.931 & 0.752 & 0.995 & 0.738 \\
    & 336 & \textbf{0.827} & \textbf{0.687} & 0.976 & 0.749 & 0.936 & 0.742 & 0.956 & 0.753 & 1.128 & 0.873 & 1.175 & 0.800 \\
    & 720 & \textbf{0.971} & \textbf{0.773} & 1.088 & 0.807 & 1.061 & 0.813 & 1.092 & 0.831 & 1.215 & 0.896 & 1.453 & 1.311 \\
    \midrule
\multirow{5}{*}{\centering ETTh2} 
     & 24 & \textbf{0.364} & \textbf{0.443} & 0.384 & 0.458 & 0.560 & 0.565 & 0.448 & 0.506 & 0.720 & 0.665 & 1.365 & 0.888 \\
     & 48 & 0.630 & 0.603 & \textbf{0.55} & \textbf{0.564} & 0.847 & 0.711 & 0.685 & 0.642 & 1.457 & 1.001 & 1.395 & 0.960 \\
     & 168 & 2.167 & 1.137 & \textbf{1.722} & \textbf{1.026} & 2.327 & 1.206 & 2.227 & 1.164 & 3.489 & 1.515 & 3.166 & 1.407 \\
     & 336 & 2.641 & 1.303 & \textbf{2.174} & \textbf{1.193} & 2.665 & 1.324 & 2.803 & 1.360 & 2.723 & 1.340 & 3.256 & 1.481 \\
     & 720 & \textbf{2.623} & \textbf{1.349} & 2.642 & 1.383 & 2.690 & 1.365 & 2.849 & 1.436 & 3.467 & 1.473 & 3.690 & 1.588 \\
     \midrule
\multirow{5}{*}{\centering ETTm1} 
     & 24 & 0.370 & 0.398 & 0.453 & 0.445 & 0.417 & 0.420 & 0.438 & 0.435 & \textbf{0.323} & \textbf{0.369} & 0.324 & 0.374 \\
     & 48 & 0.485 & 0.472 & 0.604 & 0.523 & 0.526 & 0.484 & 0.582 & 0.555 & 0.494 & 0.505 & \textbf{0.477} & \textbf{0.450} \\
     & 96 & \textbf{0.526} & \textbf{0.501} & 0.622 & 0.537 & 0.573 & 0.516 & 0.602 & 0.537 & 0.678 & 0.614 & 0.636 & 0.602 \\
     & 288 & \textbf{0.590} & \textbf{0.551} & 0.686 & 0.586 & 0.648 & 0.577 & 0.709 & 0.610 & 1.056 & 0.786 & 1.270 & 1.351 \\
     & 672 & \textbf{0.684} & \textbf{0.612} & 0.774 & 0.644 & 0.758 & 0.649 & 0.826 & 0.687 & 1.192 & 0.926 & 1.381 & 1.467 \\
     \midrule
\multirow{5}{*}{\centering ETTm2} 
     & 24 & \textbf{0.129} & \textbf{0.244} & 0.173 & 0.293 & 0.172 & 0.293 & 0.189 & 0.310 & 0.147 & 0.277 & 1.452 & 1.938 \\
     & 48 & \textbf{0.189} & \textbf{0.304} & 0.253 & 0.362 & 0.263 & 0.377 & 0.256 & 0.369 & 0.267 & 0.389 & 2.181 & 0.839 \\
     & 96 & \textbf{0.270} & \textbf{0.375} & 0.371 & 0.446 & 0.397 & 0.470 & 0.402 & 0.471 & 0.317 & 0.411 & 3.921 & 1.714 \\
     & 288 & 0.783 & \textbf{0.656} & \textbf{0.728} & 0.662 & 0.897 & 0.733 & 0.879 & 0.724 & 1.147 & 0.834 & 3.649 & 3.245 \\
     & 672 & 3.053 & 1.328 & \textbf{1.702} & \textbf{1.144} & 2.185 & 1.144 & 2.193 & 1.159 & 3.989 & 1.598 & 6.973 & 1.719 \\
     \midrule
{Avg. Rank} & & \textbf{1.6} & \textbf{1.55} & 3.0 & 2.8 & 2.85 & 2.9 & 3.7 & 3.86 & 4.45 & 4.65 & 5.4 & 5.25 \\
{Ranks 1st} & & \textbf{11} & \textbf{12} & 5 & 4 & 2 & 2 & 0 & 0 & 1 & 1 & 1 & 1 \\
    
\bottomrule
\end{tabular}
}  
\caption{Forecasting results on ETT datasets across multiple horizons.}
\end{table}

\end{document}